%% file: Main.tex
\documentclass[letterpaper, 10 pt, conference]{ieeeconf}
\IEEEoverridecommandlockouts		
\usepackage{balance}

\usepackage{amsmath}
\usepackage{amssymb}
\usepackage{pifont}% http://ctan.org/pkg/pifont

\usepackage{tabularx}
\usepackage{graphicx}
\usepackage{bm}
\usepackage{color}
\usepackage{float}
\usepackage{units}
\usepackage{comment}
\usepackage{hyperref}

\makeatletter 
\def\endfigure{\end@float} 
\def\endtable{\end@float}
\makeatother 

\usepackage{subfig}

\newcommand{\update}[1]{\textcolor{black}{#1}}

\newtheorem{remark}{Remark}

\input{Utils/Commands.tex}
\input{Utils/pmw_commands.tex}

\IEEEoverridecommandlockouts

\begin{document} 

% Paper title (needs to change)
\title{\Large \bf 			
% 	A framework for 3D optimized jumping on quadruped robots: contact timings and whole-body trajectory optimization
Contact-timing and Trajectory Optimization for 3D Jumping on Quadruped Robots
	\\[-1ex]
}
	
%\newtheorem{assumption}{Assumption}
%\begin{comment}
\author{Chuong Nguyen and Quan Nguyen 
\thanks{This work is supported by USC Viterbi School of Engineering startup funds.}
\thanks{The authors are with the Department of Aerospace and Mechanical Engineering, University of Southern California, USA: {\tt\small vanchuong.nguyen@usc.edu, quann@usc.edu, }}%
}%
%\end{comment}
												
% make the title area
\maketitle

\begin{abstract}
Performing highly agile acrobatic motions with a long flight phase requires perfect timing, high accuracy, and coordination of the full-body motion. To address these challenges, we present a novel 
approach on
timings and trajectory optimization framework for legged robots performing aggressive 3D jumping. In our 
method, we firstly utilize an effective optimization framework using simplified rigid body dynamics to solve for contact timings and a reference trajectory of the robot body. The solution of this module is then used to formulate a full-body trajectory optimization based on the full nonlinear dynamics of the robot. This combination allows us to effectively optimize for contact timings while ensuring that the jumping trajectory can be effectively realized in the robot hardware. We first validate the efficiency of the proposed framework on the A1 robot model for various 3D jumping tasks such as double-backflips off the high altitude of 2m. Experimental validation was then successfully conducted for various aggressive 3D jumping motions such as diagonal jumps, barrel roll, and double barrel roll from a box of heights 0.4m and 0.9m, respectively.

\end{abstract}

% Introduction and literature review
\input{Sections/Introduction.tex}

\input{Sections/RobotModel.tex}

% Main theory behind the adaptive support polygon

\input{Sections/TrajectoryOptimization}

\input{Sections/JumpingControl.tex}

%\input{Sections/LandingControl.tex}

% Simulation / experimental results

\input{Sections/Results.tex}

% Conlusion
\input{Sections/Conclusion.tex}
% \newpage
\balance
\bibliographystyle{ieeetr}
\bibliography{reference}
% Document end
\end{document}

%% file: Utils/pmw_commands.tex
\usepackage{multirow}
{\end{tabbing} \end{mdframed} \vspace{5px}}

\newcommand{\BM}{\begin{bmatrix}}
\newcommand{\EM}{\end{bmatrix}}
\newcommand{\beq}{\begin{equation}}
\newcommand{\eeq}{\end{equation} }

%% file: Sections/Introduction.tex
%!TEX root = ../Main.tex

\section{Introduction}
\label{sec:Introduction}
In the last decade, there has been a rapid development of legged robots, allowing them to effectively traverse rough terrain.
%to traverse rough terrain.
% (e.g.,\cite{QuannCDC16},\cite{QuannRSS17},\cite{QuannAFR2020},\cite{Yucai19},\cite{Raibert08})
% \cite{QuannRSS17},\cite{Yucai19},\cite{Raibert08}. 
Especially, the realization of jumping behaviors on legged robots has greatly drawn research attention because of their advantages in navigating high obstacles \cite{Yanran17},\cite{HWPark2021},\cite{matthew_mit2021_1},\cite{QuannICRA19}.

% With recent advancement of hardware and well-established control strategies, quadruped robots has shown its impressive capabilities to perform various dynamic locomotion
% \cite{GerardoBledt},\cite{Hutter16},\cite{Donghyun19}.
% Specifically regarding to quadruped jumping behaviors, the Cheetah 2 recently demonstrates jumping autonomously over obstacles while bounding at high-speed \cite{HWPark2021}, and the Cheetah 3 is able to repeatably jump onto and jump down from a desk up to $30''$ in height \cite{QuannICRA19}.

% The dynamics of quadruped robots is highly complexity: highly nonlinear, high DOF (typically $18$) and multiple contact models with the ground. This makes the quadruped robots difficult to control. 
% In addition, in our TO setup, we use the cost function to minimize the GRF related to the energy, while finding the contact timings. This guarantees the successful jumps on hardware due to the limited torque of the quadruped robots.

\begin{figure}[!t]
	\centering
	{\centering
		\resizebox{0.8\linewidth}{!}{\includegraphics[trim={0cm 14cm 0cm 8cm},clip]{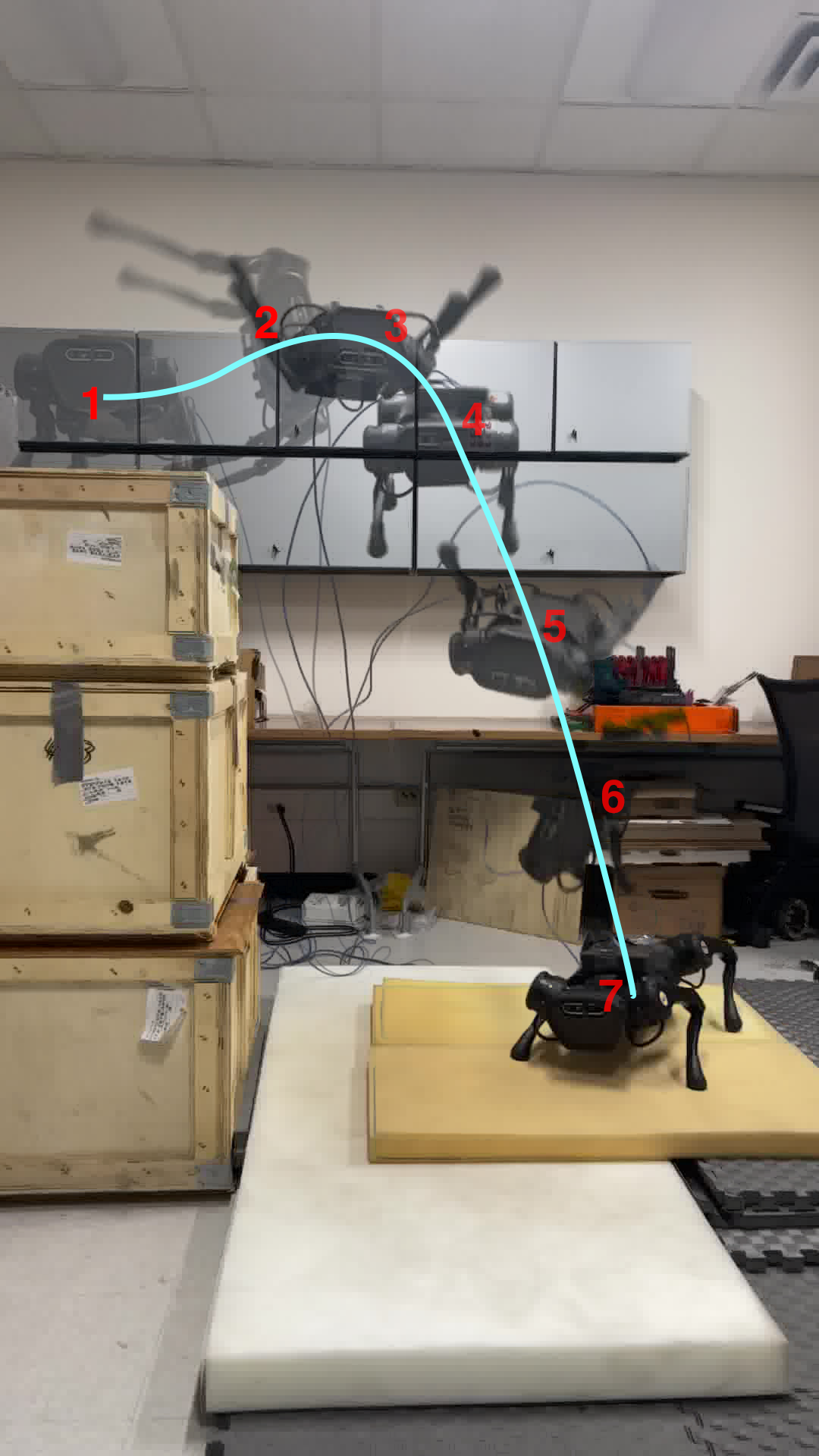}}
	}
	\caption{
	A combined motion snapshot of the Unitree A1 robot performing a double barrel roll from a $0.9m$ platform. Experiment and simulation video: \url{https://youtu.be/d7RcWEXTbqc}.}
	\label{fig:double_roll}
\end{figure}

There exists a number of dynamic models of legged robots, which are commonly utilized by optimization and control frameworks to realize agile locomotion. 
% \todo{This sentence alone does not make sense. the one you have commented out is better to start the statement.}
Single rigid body dynamics (SRBD) is widely used for control of legged robots. This simplified model considers the effect of ground reaction forces on the robot body while ignoring the leg dynamics, which enables real-time computation, e.g., Model Predictive Control (MPC), to achieve highly dynamic motions in \cite{HWPark2021},\cite{Carlo2018},\cite{YanranDingTRO},\cite{GabrielICRA2021}. However, real-time MPC typically requires a limited prediction horizon when planning these motions.
% Our method instead focuses on addressing acrobatic motions with a long fight phase}.
% This simplification enables real-time feedback control. This model can also be utilized to achieve 3D jumping via Model Predictive Control (MPC) \cite{YanranDing19}. 
% However, to guarantee real-time computation, the prediction horizon of the MPC controller is limited, limiting the approach from addressing motions with a long aerial time.  
% Since the legs' dynamics is neglected, this simplification hardly achieve accurate  affects the accuracy of jumping motions .
% However, they sacrifice the accuracy for online execution when motions with long aerial time are executed. 
% \update{Based on this simplified model, the authors in \cite{YanranDingTRO} and \cite{GabrielICRA2021} propose Model Predictive Control (MPC) to achieve agile and acrobatic motions. 
% However, their work assumes the limited prediction horizon to guarantee real-time computation, thus hardly achieving accurate motions with a long aerial time.}
% \todo{Don't try to complain other people's work directly. You can try sth like "However, real-time MPC typically will require ... . Our approach instead focuses on addressing ... with a long fight phase."}
While the SRBD model is a suitable choice
for online execution, the full-body dynamics can be adopted when the model accuracy is crucial \cite{QuannICRA19},\cite{Posa14}. 
% The FBD is utilized in trajectory optimization (TO) framework to accurately plan a bipedal robot's running gait in a planar plane \cite{Posa14}. 
Acrobatic jumping motions usually require a long aerial phase and high accuracy of full-body coordination for all contact phases. Thus, it is essential to consider the full-body dynamics. 
In our prior work \cite{QuannICRA19}, we have introduced an effective trajectory optimization framework that allows MIT Cheetah 3 jumping on a high platform. However, this framework is primarily designed for 2D motion with a predefined contact schedule. In this paper, we present a unified framework to optimize for contact schedule and 3D jumps on legged robots.  
% \update{In this paper, we extend our previous work  
% that implements jumps in a planar plane to achieve highly complex and accurate 3D jumps with the full-body dynamics model.}

% \update{This inspires us to use the full-body dynamics in a trajectory optimization framework to accurately plan and execute highly agile jumps in 3D} 
% Our previous work \cite{QuannICRA19} employs a 2D quadrupedal model to achieve accurate jumping motions in a planar plane. 
% \update{In this paper, we extend our previous work \cite{QuannICRA19}, which considers 2D jumps, to achieve highly complex jumping motions in 3D.}
% Regarding other type of motions, the TO of full rigid-body dynamics is utilized to accurately plan periodic running gait constrained in 2D planar plane for a 13 DoF bipedal robot in \cite{Posa14}, and for a rearing and posture recovery in place in \cite{andreea_ICRA17}.

Centroidal dynamics and full-body kinematics constraints are also used in trajectory optimization to execute agile motions \cite{matthew_mit2021_1},\cite{Dai14}. They reduces the complexity of dynamics compared to the full-body dynamics, while considering the feasibility of kinematic motions. 
% Thus, motion planners are still able to achieve some 3D motions while keeping the computational tractability. 
Regarding the
quadruped jumping, a recent work in \cite{matthew_mit2021_1} combines trajectory optimization using the centroidal dynamics and a whole-body controller to track the reference motions from the optimization.
% \todo{However, one limitation of this framework is that torque ($\bm{\tau}$) is not constrained directly in the TO but indirectly via GRF ($\mathbf{F}$) based on the approximation $\bm{\tau}=\mathbf{J}^T \mathbf{F}$, where the Jacobian $\mathbf{J}$ is also approximated. This approximation is likely to cause accumulated error in jumping trajectory over time, especially for jumping tasks with long aerial phase}. 
Difference from \cite{matthew_mit2021_1}, we optimize contact timings and leverage the full nonlinear dynamics of the robot, which allows us to achieve highly accurate jumps such as multiple backflips and barrel rolls (see Fig. \ref{fig:double_roll}).

While there is an increasing number of work attempts to find optimal contact timings for agile motions \cite{Igor2012},\cite{Neunert17},\cite{Winkler_RAL2018},\cite{Ponton16}, the contact timings optimization for highly dynamic 3D jumps has not been well explored.
The recent work \cite{Winkler_RAL2018} proposes a trajectory optimization (TO) approach based on SRBD to solve for contact timings. However, since this work uses a linearized rigid body dynamics
using Euler angle representation, it offers a limited range of achievable motions due to the singularity \cite{TaeyoungLee}.
In our work, we are interested in optimizing the contact timings for highly aggressive and complex 3D jumps with long aerial time. We utilize a rotation matrix to represent body
orientation to prevent the singularity and unwinding issues associated with Euler angle and quaternion
representation \cite{TaeyoungLee}. 
% \textcolor{blue}{This allows us to plan complex jumping motions in 3D while optimizing the contact timings.}  
In addition, \textcolor{black}{difference from \cite{Winkler_RAL2018}} , the solution of our proposed contact-timing optimization is then utilized to formulate the full-body TO that considers torque constraints and the whole-body coordination. Therefore, our framework enables aggressive jumps to be realized in the robot hardware.

% In addition, in our TO setup, we use the cost function to minimize the GRF related to the energy, while finding the contact timings. This guarantees the successful jumps on hardware due to the limited torque of the quadruped robots.

% Our work finds most relevant with impressive work \cite{matthew_mit2021_1} \cite{YanranDing19},\cite{Winkler_RAL2018}, which propose some approaches based on simplified dynamics to achieve  a number of complex jumping motions in quadruped robots.

% that crucially requires the accurate model (e.g. heavy-leg quadruped robot), the whole body dynamics model has remarkable advantage over the SRB dynamics model. 
% Our previous work \cite{QuannICRA19} simplifies the whole-body dynamics into 2D with $7$ degree of freedom, then implements successfully the jumping on the MIT Cheetah 3 robot. 
% Extending this to 3D jumping behaviors is challenging due to the complexity of the model, a highly nonlinear dynamic constrains, and a large number of other constraints related to joints' kinematics, torque, configuration and obstacles clearance. 

% In addition, we propose an TO-based approach to automatically compute the optimal contact timings, which is then used for the whole-body TO. 

The contribution of the work is summarized as follows. 
Firstly, we propose a contact-timing optimization method to simultaneously solve for optimal contact timings and reference trajectory of the robot body, which will then be used to formulate a full-body trajectory optimization. 
Secondly,
in the contact-timing optimization, the rotation matrix is directly utilized to represent the orientation of the robot body to avoid singularity and unwinding issues. This utilization allows us to optimize for a wide range of complex 3D jumping motions. We also present a full-body trajectory optimization to perform a diverse set of aggressive 3D jumping tasks that require high accuracy and long aerial time. Finally, our proposed framework is validated in both experiment and the A1 robot model for various gymnastic 3D jumping tasks, e.g., double-backflips and double barrel roll from boxes of height 2m and 0.9m.

The rest of our paper is organized as follows. The optimization framework are presented in Section \ref{sec:optimization}; jumping and landing controllers are briefly described in Section \ref{sec:JumpingControl}. Results from selected hardware experiments are presented in Section~\ref{sec:Results}. Finally, Section~\ref{sec:Conclusion} provides concluding remarks.

%% file: Sections/RobotModel.tex
%!TEX root = ../Main.tex

% \section{Hardware Platform For Verification}
% \label{sec:RobotModel}

% 	\begin{figure}[t!]
% 	\hspace{0.5cm}
% 	\center
% 	\includegraphics[width= 0.9\columnwidth]{Figures/A1_robot}%RobotStairs-crop.pdf}%
% 	\caption{{\bfseries LASER robot.} Each leg consists of three actuators.}
% 	\label{fig:A1_robot}
%     \end{figure}

% \begin{figure}[!t]
% 	\centering
% 	{\centering
% 		\resizebox{1\linewidth}{!}{\includegraphics[trim={0cm 4cm 0cm 2cm},clip]{Figures/A1_robot}}
% 	}
% 	\\
% 	\caption{\textbf{A1 robot.} Each leg consists of three actuators.} 
% 	\label{fig:A1_robot}
% \end{figure}

%% file: Sections/TrajectoryOptimization.tex
\section{Contact-timing optimization}
\label{sec:optimization}

\begin{comment}
\begin{figure}[!t]
	\centering
	\resizebox{0.8\linewidth}{!}{\includegraphics[trim={2cm 3.7cm 2cm 4cm},clip]{Figures/3D_DynamicsModel.pdf}}
	\caption{{\bfseries 3D Quadruped Model.} 
%		Definition of system states and control inputs for the dynamics presented in Section \ref{subsec:dynamics}. 
		The jumping trajectory optimization employs a simplified 3D model of A1 robot. This figure shows the choice of coordinates and the control inputs of the simplified 3D model.}
	\label{fig:model}
\end{figure}
\end{comment}

\subsection{Motivation}
When performing highly dynamic and complex jumping
in 3D, it is critical to leverage the full-body dynamics of the robot to maximize the jumping performance and guarantee the high accuracy of the jumping trajectory while realizing in the real
robot hardware.
% To the best our knowledge, it does not have any work on implementing the full-body TO to achieve 3D quadruped jumping motions.

In the full-body trajectory optimization, contact timing is predefined for each jumping phase \cite{QuannICRA19}. 
During the flight phase of the jumping motion, the robot motion has minimal effect in the COM trajectory of the robot. Therefore, it is critical to optimize for the contact schedule, especially with motions that have a significant aerial time.
% Since jumping motions usually have long aerial phase and there are no control inputs to change the center of mass (CoM) trajectory of the robot during that phase, optimizing the contact timings is essential for the entire motion. 
Moreover, the manual selection of the timings is time-consuming, not optimal, and even it is not feasible for the full-body trajectory optimization to obtain solutions for many complex 3D jumps. Therefore, it is crucially important to implement an approach to automatically compute optimal timings. Furthermore, timings in highly dynamic motion with long flight phase plays a crucial role in minimizing the effort or energy, guaranteeing the feasibility of the motion within the limits of actuator powers.

In the following, we introduce a framework that includes
optimal contact timings and full-body trajectory optimization to generate complex 3D jumping motions at high accuracy.

\subsection{Contact timings optimization} \label{subsec:timing_optimization}

\begin{figure}[!t]
	\centering
	{\centering
		\resizebox{1.1\linewidth}{!}{\includegraphics[trim={0cm 5cm 0cm 0cm},clip]{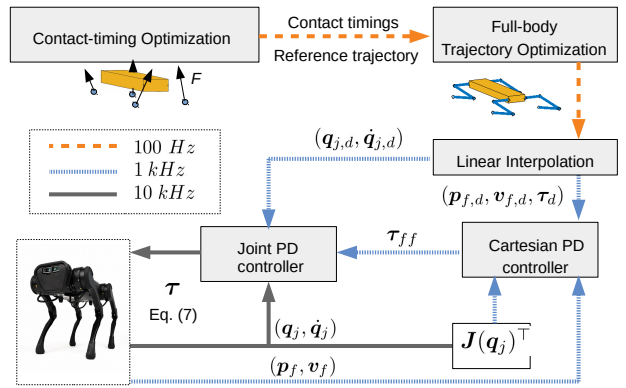}}
	}
	\\
	\caption{Block diagram of the propose framework. 
% 	The contact-timing optimization using SRBD produces the optimal contact schedule and reference trajectory at $100~Hz$, which then be used for the full-body TO. Linear interpolation is then used to get the reference profile at $1~kHz$ for the Cartesian controller. The joint PD controller executes at $10~kHz$.
	} 
	\label{fig:control_diagram}
\end{figure}

A direct implementation of contact-timing optimization using the full nonlinear dynamics of the robot takes considerable time to solve due to the high complexity of the problem. In our implementation, it does not even produce a feasible solution for many complex 3D jumps. 
Therefore, these issues motivate us to take advantage of the simplified dynamics to optimize for contact timings. Section \ref{subsec:timing_optimization} presents our contact-timing optimization framework to obtain the optimal contact schedule and body reference trajectory, which then be used to formulate the full-body trajectory optimization introduced in Section \ref{subsec:TO_3D}. The overview of our approach is illustrated in Fig. \ref{fig:control_diagram}.

% \begin{table}[t!]
% %	\hspace{0.2cm}
% 	\centering
% 	\caption{A1 Robot Parameters}
% 	\label{tab:PRP}
% 	\begin{tabular}{cccc}
% 		\hline
% 		Parameter & Symbol & Value & Units\\
% 		\hline
% 		Gear Ratio & $gr$    & 9 & $\unit{}$  \\[.5ex]
% 		Max Torque & $\tau_{max}$    & 33.5 & $\unit{Nm}$  \\[.5ex]
% 		Max Joint Speed & $\dot{q}_{max}$    & 21 & $\unit{rad}/\unit{s}$  \\[.5ex]
% 		Total leg mass & $m_l$    & 4.71 & $\unit{kg}$  \\[.5ex]
% 		Total robot mass & $m$    & 12 & $\unit{kg}$  \\[.5ex]
% 		Trunk dimension & $l,w,h$ & 0.361, 0.194, 0.114 & $\unit{m}$ \\[.5ex]
% 		Trunk Inertia  & $I_{xx},I_{yy},I_{zz}$  & 0.017, 0.056, 0.065 & $\unit{kg}.\unit{m}^2$ \\[.5ex]
% 		Hip Link Lengths & $l_{1}$ & 0.083 & $\unit{m}$ \\[.5ex]
% 		Leg Link Lengths & $l_{2}, l_{3}$ & 0.2 & $\unit{m}$ \\[.5ex]
% 		\hline 
% 	\end{tabular}
%     \end{table}

\begin{table}[t!]
%	\hspace{0.2cm}
	\centering
	\caption{Optimal contact timings $T_i$, solving time via \textit{IPOPT} for the contact-timing optimization in Section \ref{subsec:timing_optimization}, and full-body trajectory optimization in Section \ref{subsec:TO_3D}}
	\begin{tabular}{cccc}
		\hline
		Jumps & Optimal timings  & Solving time & Solving time\\
		 & $T_i$ ($\times 10ms$) & contact-timing  & full-body\\
		&  & optimization & TO\\
		\hline
		lateral 30cm & 50, 28    & 4.14 [s] & 31 [s] \\[.5ex]
		lateral down & 52, 35   & 5.02 [s] & 32 [s] \\[.5ex]
		$90^{0}$ spinning & 56, 31    & 5.7 [s] & 39 [s] \\[.5ex]
		diagonal jump & 54, 30, 33   & 5.98 [s] & 226 [s] \\[.5ex]
		barrel roll & 51, 32, 35   & 6.61 [s] & 35 [s] \\[.5ex]
		double barrel roll & 52, 34, 55    &  11.7 [s] & 46 [s] \\[.5ex]
		double backflip & 50, 33, 69   & 7.23 [s] & 89 [s] \\[.5ex]
		\hline 
		\label{tab:TO_solving_time}
	\end{tabular}
\end{table}

 Unlike \cite{matthew_mit2021_1} and \cite{Winkler_RAL2018}, which use Euler angles, we use a rotation matrix to represent the body orientation when planning 3D jumping motions. 
%  Our utilization prevents the singularity issue associated with Euler angles, and also avoids the unwinding issue related with quaternions representation \cite{TaeyoungLee}. Thus, our framework allows to generate any 3D jumping motions.
%  \cite{YanranDing19},\cite{Shuster1993},\cite{Sanjay2000}. 
% However, using rotation matrix in the TO set up introduces more optimization variables and constraints, which causes significantly solving time. In many cases, it cannot provide a feasible solution. Thus, it is challenging and hinders the use of rotation matrix in the TO set up so far. To tackle this challenge, we propose a Taylor series approximation at a high degree to approximate the rotation matrix in $SO(3)$ and consider an error term of rotation matrix.
Given the sequence of contacts, we will optimize their duration (i.e., contact timings). We choose Cartesian space to derive the SRB's equation of motion:
\begin{subequations}\label{eq:SRB_dynamics}
    \begin{align}
        \ddot{\bm{p}}&=\sum_{s=1}^{n_s} \bm{f}^{s}/m-\bm{g}, \label{eq:SRB_dynamics1}\\
        \bm{I_b} \dot{\bm{\Omega}} + \bm{\Omega} \times \bm{I_b} \bm{\Omega} &= \bm{R}^{-1} \sum_{s=1}^{n_s} \bm{f}^s \times (\bm{p} - \bm{p}_f^s) ,\label{eq:SRB_dynamics2}\\
        \dot{\bm{R}} &= \bm{R}\widehat{\mathbf{\Omega}}\label{eq:SRB_dynamics3}
    \end{align}
\end{subequations}

where $n_s$ is number of feet, m is the robot's mass, $\bm{\Omega} \in \mathbb{R}^3$ is angular velocity expressed in the body frame, $\bm{R}$ is rotation matrix of the body frame, $\bm{g}$ is the gravity acceleration; $\bm{p}, \dot{\bm{p}}, \ddot{\bm{p}} \in \mathbb{R}^3$ is the CoM position, velocity, acceleration of the body in the world frame; $\bm{f}^s \in \mathbb{R}^3$ is GRF on foot $s^{th}$;   $\bm{p}_f^s \in \mathbb{R}^3$ is $s^{th}$ foot position in the world frame. The hat map $(\hat{.}):\mathbb{R}^3  \rightarrow so(3) $ converts any vector in $\mathbb{R}^3$ to the space of skew-symmetric matrices. For the sake of notation, we define the robot's state as
$\bm{x}:=\left[\bm{p}; \dot{\bm{p}}; \bm{\Omega}; \dot{\bm{\Omega}}; \bm{R}\right]$. The contact-timing optimization is then formulated as follows:
\begin{subequations} \label{eq:time_opt}
\begin{align}
&\underset{\bm{x}, \bm{f}, \bm{e}_R}{\text{minimize}} \textrm{ } \sum_{k=1}^{N^c}\epsilon_{\Omega} \bm{\Omega}_k^T\bm{\Omega}_k+\epsilon_{f}\bm{f}_k^T\bm{f}_k + \textcolor{black}{\epsilon_R \bm{e}_{R_k}^T \bm{e}_{R_k}}\nonumber\\
\text{s.t. }
             & [\bm{R},\bm{p},\bm{p}_f^s](k=1)=[\bm{R}_0,\bm{p}_0,\bm{p}_{f,0}^{s}], ~ \textrm{\textit{initial states}} \label{eq:time_opt_1} \\ 
             & [\bm{\Omega},\dot{\bm{\Omega}},\dot{\bm{p}}](k=1)=\bm{0}, ~~~~~~~~~~~~~~~~\textrm{\textit{initial states}} \label{eq:time_opt_1_2}   \\         
             & [\bm{R},\bm{p},\bm{p}_f^s](k=N^c)=[\bm{R}_g,\bm{p}_g,\bm{p}_{f,g}^{s}],\textrm{\textit{final states}} \label{eq:time_opt_2} \\ 
            %  & \mathbf{p}_f^s(k) \in \mathcal{B}_s(b,\mathbf{R}(k),\mathbf{p}(k)), ~~~~~~~~~~\textrm{\textit{foot position}} \label{eq:time_opt_3}\\
             & |\bm{R}(k)[\bm{p}_f^s(k)-\bm{p}(k)]- \bar{\bm{p}}_f^s(k)| \leq \textbf{r}, \textrm{\textit{foot position}} \label{eq:time_opt_3}\\
             & \update{|\bm{f}_k^{s,x}/\bm{f}_k^{s,z}|\le \mu, |\bm{f}_k^{s,y}/\bm{f}_k^{s,z}|\le \mu, ~~\textrm{\textit{friction cone}}} \label{eq:time_opt_4_1} \\ 
             & \update{\bm{f}_{k, min}^s \leq \bm{f}_k^{s,z} \le \bm{f}_{k, max}^s, ~~~~~~~~~~~~~\textrm{\textit{GRF limits}}} \label{eq:time_opt_4_2} \\
             & \update{\bm{p}_{k,min} \leq \bm{p}_k \leq \bm{p}_{k,max},~ \textrm{\textit{CoM in contact phases}}}  \label{eq:time_opt_4_3} \\
             & \gamma(\bm{x}_k, \bm{x}_{k+1}, \bm{f}(k), \bm{p}_f^s(k)) =0,~\textrm{\textit{discrete dynamic}} \label{eq:time_opt_5} \\
             & \textcolor{black}{\bm{R}_{k+1} = \bm{R}_{k}exp(\widehat{\bm{\Omega}_{k} T_i/N_i})}, ~~~~~ \textit{SO(3) manifold}  \label{eq:time_opt_6}\\
             & \sum_{i=1}^{n_p} T_i  \in [T_{min}, T_{max}] , ~ N^c = \sum_{i=1}^{n_p} N_i  \label{eq:time_opt_7}\\
             & \textrm{ for } \textrm{ }  k=1,2,... N^c \nonumber
\end{align}
\label{eq:timings_TO_of_SRB}
\end{subequations}
where $\bm{\Omega}_k \in \mathbb{R}^{3}$, $ \bm{f}_k \in \mathbb{R}^{12}$ is angular velocity of SRB w.r.t the body frame, and GRF on four legs at the iteration $k^{th}$; $\epsilon_{\Omega}, \epsilon_{f}, \epsilon_R $ are cost function weights of corresponding elements, and $\epsilon_R, \epsilon_{\Omega} >> \epsilon_{f}$. We use an error term of rotation matrix as $\bm{e}_{R_k}= log(\bm{R}_{ref,k}^T . \bm{R}_k )^{\vee}$, where $log(.): SO(3) \rightarrow so(3)$ is the logarithm map, and the vee map $(.)^{\vee}: so(3) \rightarrow \mathbb{R}^3$ is the inverse of hat map \cite{TaeyoungLee},\cite{Bullo_GC04}. With given final and initial rotation matrix, $\bm{R}_g$ and $\bm{R}_0$ respectively, we utilize a linear interpolation to obtain $\bm{R}_{ref,k}$ at $k^{th}$ step.
The equation (\ref{eq:time_opt_3}) implies that the $s^{th}$ foot position is constrained inside a sphere $\mathcal{S}_s$ of radius $\textbf{r}$ so that the joint angle are within limits. The center of the sphere $\bar{\bm{p}}_f^s(k)$ is relative to the CoM position.
% The function $\beta(.)$ captures
% various constraints on CoM, friction cone limits,
% GRF, and geometric constraints related to the ground and obstacle clearance. 
The function $\gamma(.)$ captures the dynamic constraints discretized from (\ref{eq:SRB_dynamics1})-(\ref{eq:SRB_dynamics2}) via the forward Euler method.

In (\ref{eq:time_opt_7}), $n_p$ is a number of contact phases. For example, if the pre-selected contact schedule is four-leg contact, rear-leg contact, and flight phase, then $n_p=3$. $N_i$ is the predefined number of time steps for the $i^{th}$ contact phase, $i \in \left\{1,2,..,n_p \right\}$. 
Note that our approach solves for optimal timing $T_i$, given a total predefined interval $[T_{min}, T_{max}]$. The equation (\ref{eq:time_opt_6}) is derived from (\ref{eq:SRB_dynamics3}) to ensure $\bm{R}_k$ evolves in the SO(3) manifold. Here, $exp(.):so(3) \xrightarrow[]{} SO(3)$ is known as the matrix exponential map.

% \textcolor{blue}{The algorithm will check a $k^{th}$ iteration belongs to which $i^{th}$ predefined contact phase to choose $T_i/N_i$ accordingly.} 

% The term $\bm{f}_k^T\bm{f}_k$, which is intuitively represented for generated power (\cite{matthew_mit2021_1},\cite{YanranDing19}), is added to the cost function of (\ref{eq:timings_TO_of_SRB}). By utilizing that term, we are able to minimize necessary effort or energy of robot when perform jumping tasks.
\begin{remark}
Despite the advantage of using a rotation matrix $\bm{R}$ to achieve the most 3D jumping motions, utilizing it in the trajectory optimization setup introduces more optimization variables and constraints. This significantly increases the problem size and solving time.
To achieve a feasible solution, $\bm{R}$ in (\ref{eq:time_opt_6}) must be approximated at high accuracy enough to satisfy the $SO(3)$ property at every time step. Taylor series approximation at a high degree is utilized for that purpose. However, choosing a really high degree of Taylor's approximation is prohibitively costly in terms of computational time. Therefore, in order to balance accuracy and solving time, we utilize the $4^{th}$ degree of Taylor's series:
$exp(\mathbf{A})= \sum_{k=0}^4 \frac{\mathbf{A}^k}{k!}.$
Moreover, for complex jumping motions in 3D, adding and minimizing the term $\bm{e}_{R_k}^T \bm{e}_{R_k}$ in the cost function in (\ref{eq:time_opt}) helps to guide the trajectory optimization toward a feasible solution in a fast manner.
\end{remark}
% \begin{align}
%     exp(\mathbf{A})=\mathbf{I}+\mathbf{A}+ \frac{1}{2} \mathbf{A}^2 + \frac{1}{6} \mathbf{A}^3 + \frac{1}{24} \mathbf{A}^4.
% \end{align}
% \begin{remark}
% Our method for optimal contact timing also gives an advantage that it is general to jumping motions of legged robots, \update{which their dynamics can be simplified as SRBD},  with arbitrary numbers of legs and long aerial time.
% \end{remark}

% \begin{remark}
% Different from \cite{Winkler_RAL2018}, regarding to the finding of contact timings, we use a cost function to guide the TO, and use rotation matrix instead of euler angles. Since we specifically aim to perform dynamic jumping tasks, the cost function plays a key role. This helps minimize the power the motor generated to tackle torque saturation issues, while computing optimal contact timings respectively. \todo{This remark is not really important. You can remove this.}
% \end{remark}

\subsection{Full-body trajectory optimization for 3D jumping} \label{subsec:TO_3D}
When performing highly dynamic jumping, it is important to consider the full nonlinear dynamics of the robot in the optimization framework. This will guarantee the accuracy of the jumping trajectory while transferring to the hardware.
% \textcolor{blue}{When performing highly dynamic jumping, it is important to enforce directly full nonlinear dynamics constraints at every optimization step to achieve high accuracy of the jumping trajectory.
% that crucially requires the accurate model (e.g. heavy-leg quadruped robot), using the whole-body TO is very important to achieve highly accurate jump. 
% For example, in the problem of performing barrel or back-flip off a high elevation, it may not be feasible for the robot to recover from unexpected landing configurations and the hardware is able to be damaged due to the hard impact at these configurations. 
% Therefore, the role of full-body TO is crucially important to ensure a small error in the robot configuration at the end of the jumping phase}

\subsubsection{Full-body dynamics model}
\label{subsubsec:3Dmodel}
\label{sec:JumpingControl}

\begin{figure}[!t]
	\centering
	{\centering
		\resizebox{1\linewidth}{!}{\includegraphics[trim={0cm 0cm 0cm 0cm},clip]{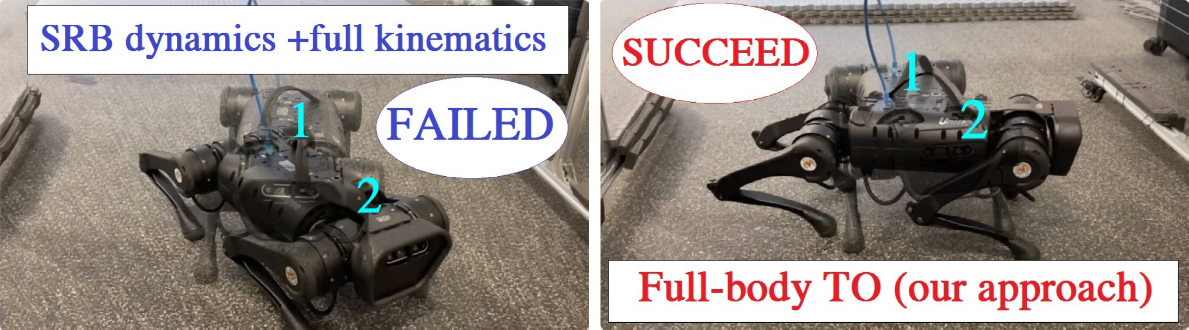}}
	}
    {\centering
		\resizebox{0.8\linewidth}{!}{\includegraphics[trim={0cm 10cm 0cm 10cm},clip]{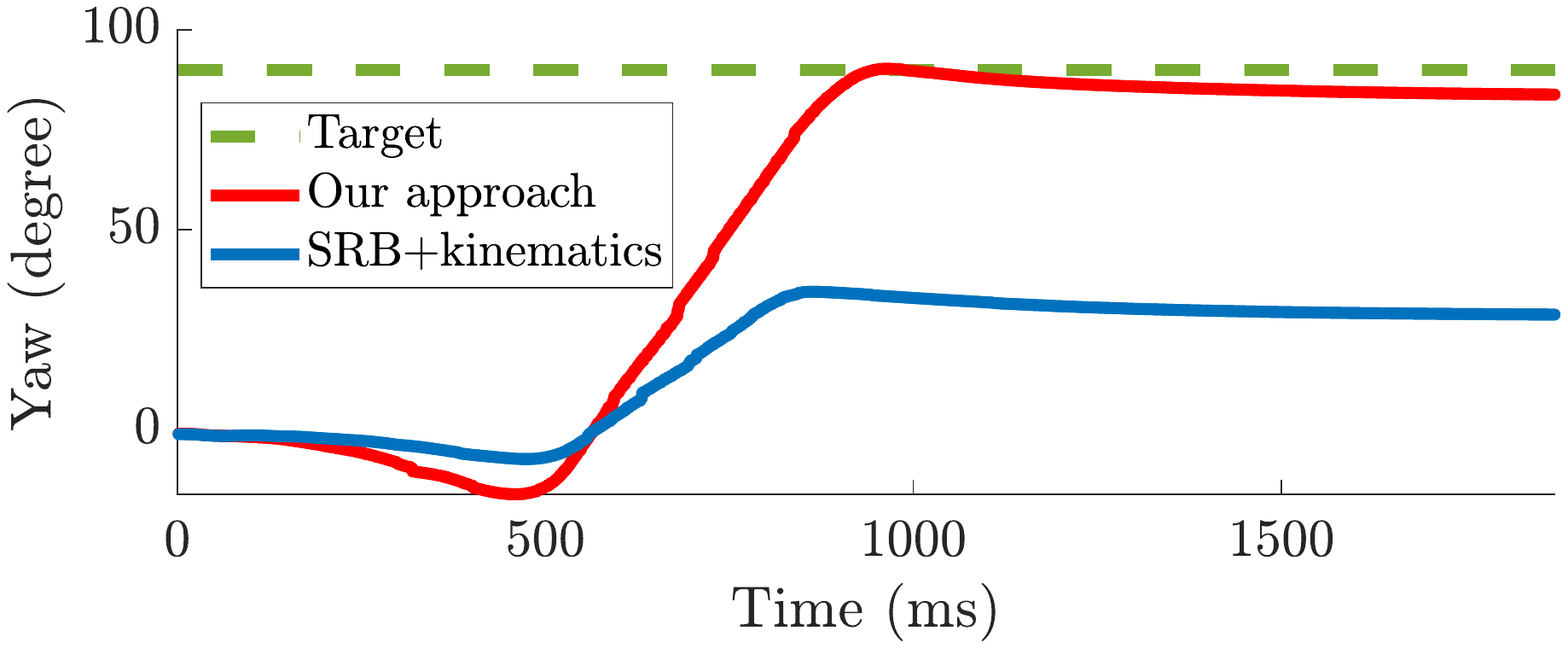}}
	}
	
	\caption{\textbf{} Comparison between \textit{trajectory optimization of SRB dynamics+ full-body kinematics constraints approach} and \textit{full-body TO (our approach)} for a $90^0$ spinning jump target in experiment.}
	\label{fig:yaw_compare}
\end{figure}

The robot is modeled as a rigid-body system, and spatial vector algebra in \cite{featherstone_rbda} is used to construct the robot's equations of motion:
%We then used the spatial vectors described in \cite{featherstone_rbda} to construct the robot model, resulting in the following dynamics:

% \begin{align}
% \label{eq:dynamics}
% & \bm{H}(\bm{q})\bm{\ddot{q}}+\bm{C}(\bm{q},\bm{\dot q})\bm{\dot q}+\bm{g}(\bm{q})\nonumber\\
% & =\bm{B}\bm{\tau}+\bm{B}_{fric}\bm{\tau}_{fric}(\bm{\dot q})+\sum_{s} \bm{J}_s^T(\bm{q})\bm{F}_s,
% \end{align}

\begin{equation}\label{eq:full_dynamics}
\begin{bmatrix}\bm{H} & -\bm{J}_s^T \\ -\bm{J}_s^T & \mathbf{0} \end{bmatrix} \begin{bmatrix}
\bm{\ddot{q}} \\ \bm{F}_s
\end{bmatrix}= \begin{bmatrix}-\bm{C}\bm{\dot q}-\bm{g} + \bm{B}\bm{\tau}+\bm{B}_{f}\bm{\tau}_{f} \\ \bm{\dot J}_{s}(\bm{q})\bm{\dot q}\end{bmatrix}
\end{equation}
where $\bm{q}:=\left[\bm{q}_{b};~\bm{q}_{j}\right]$ is a vector of generalized coordinates, in which $\bm{q}_b$ represents CoM position and orientation of the robot body, and $\bm{q}_j \in \mathbb{R}^{12}$ is a vector of joint angles. The mass matrix is denoted by $\bm{H}$; the matrix $\bm{C}$ is represented for  Coriolis  and centrifugal terms; $\bm{g}$ is the gravity vector; $\bm{J}_s$ is the spatial Jacobian of the body containing the $s^{th}$ contact foot, expressed at the foot and in the world coordinate system; $\bm{B}$ and $\bm{B}_{fric}$ are distribution matrices of actuator torques $\bm{\tau}$ and the joint friction torques $\bm{\tau}_{fric}$; $\bm{F}_s$ is the spatial force at the $s^{th}$ contact foot. 

% Moreover, for each stance foot $s$ that is on the ground and actively supporting the robot, the following constraints are enforced :
% \begin{align}
% \label{eq:contact_constraints_in_dynamics}
% \bm{J}_{s,stance}(\bm{q})\bm{\ddot q}+\bm{\dot J}_{s,stance}(\bm{q})\bm{\dot q}=0.
% \end{align}
%The stance feet in contact with the ground have the following constraints:

\subsubsection{Cost function and constraints}
The ultimate goal of our framework is to find a feasible jumping motion for each 3D jumping task with the full rigid-body dynamics consideration. Due to the high complexity of this problem, the feasibility domain is limited. Therefore, the purpose of this cost function is to guide the optimization to converge to a feasible solution where the robot's coordinates stays close to the reference configuration $\bm{q}_{ref}$ if possible. The CoM position and body orientation obtained from Section \ref{subsec:timing_optimization} are also linearly interpolated to get their profiles sampling at $dt=10~ms$, which is then used as reference for the full-body TO here.
The cost function is defined as follows:
\begin{align}
J=&\sum_{h=1}^{N-1}\epsilon_q (\bm{q}_{h}-\bm{q}_{ref,h})^T(\bm{q}_{h}-\bm{q}_{ref,h})+\epsilon_{\tau}\bm{\tau}_{h}^T\bm{\tau}_{h} \nonumber\\
&+\epsilon_N(\bm{q}_{N}-\bm{q}_{N}^{d})^T(\bm{q}_{N}-\bm{q}_{N}^{d}), 
\label{eq:cost_function_wholebodyTO}
\end{align}
where $N$ denotes the total of the time steps (i.e. $N=T_{opt}/dt$ with $T_{opt}$ is optimal total time obtained from Section \ref{subsec:timing_optimization}; $\bm{q}_{h}, \bm{\tau}_{h}$ are the generalized coordinates and joint torque at the iteration $h^{th}$; $\bm{q}_{N}$ is the the generalized coordinates at the end of the trajectory; the first six elements of $\bm{q}_{ref,h}$ is the reference CoM position and body orientation obtained from contact timing optimization in Section \ref{subsec:timing_optimization}. The last 12 elements of $\bm{q}_{ref,h}$ is set to be the final joint configuration. We also use $\bm{q}_{ref,h}$ as initial guess for the TO to reduce the solving time.
$\epsilon_q, \epsilon_{\tau}, \epsilon_N$ are cost function weights of corresponding elements.
% Similarly to \cite{QuannICRA19}, we set up constraints regarding to initial and final body and joint configuration; pre-landing configuration; joint angular position and velocity limits; torque limits; friction cones; geometric constraints to ensure the body, hips, knees and swing feet have a good clearance with other robot's parts, with the ground, and with obstacle clearance.
The following constraints are:
\begin{itemize}
	\item Full-body dynamics constraints \eqref{eq:full_dynamics}
	\item Initial configuration: $\bm{q}(h=0)=\bm{q}_{0}, \bm{\dot q}(h=0)=\bm{0}$.
	\item Pre-landing configuration: $\bm{q}_{j,h}=\bm{q}_{h,N}^d, \bm{\dot q}_{j,h}=\bm{0}~ (h\in[(N-a):N])$, where $a$ is a jumping task-based selection ($a<N$).
	\item Final configuration: $\bm{q}_h(h=N)=\bm{q}_{N}^d$. 
	\item Joint angle constraints: $\bm{q}_{j,min}\le \bm{q}_{j,h} \le \bm{q}_{j,max}$
	\item Joint velocity constraints: $|\bm{\dot q}_{j,h}|\le \bm{\dot q}_{j,max}$
	\item Joint torque constraints: $|\bm{\tau}_h|\le \bm{\tau}_{max}$
	\item Friction cone limits: $|\bm{F}_h^x/\bm{F}_h^z|\le \mu$, $|\bm{F}_h^y/\bm{F}_h^z|\le \mu$
	\item Minimum GRF: $\bm{F}_{h}^z \ge \bm{F}_{min}^z$
	\item Geometric constraints to guarantee: (i) each robot part does not collide with others, (ii) the whole robot body and legs have a good clearance with obstacle.
\end{itemize}

%% file: Sections/JumpingControl.tex
\section{Jumping and Landing Controller}
\label{sec:JumpingControl}
% \begin{figure}[t!]
% 	\center
% 	\includegraphics[width=0.7 \columnwidth]{Figures/2roll/2roll_v3.jpg}%RobotStairs-crop.pdf}%
% 	\caption{{\bfseries A double barrel roll} from a box of  height $0.8m$ with our proposed framework.}
% 	\label{fig:double_barrel}
% \end{figure}

% \begin{figure}[t!]
% 	\center
% 	\includegraphics[width=0.7 \columnwidth]{Figures/backflip_2rolls/2backflip_3.jpg}%RobotStairs-crop.pdf}%
% 	\caption{{\bfseries A double backflip} from a box of height $2m$ with our proposed framework}
% 	\label{fig:double_backflip}
% \end{figure}

\begin{figure}[!t]
	\centering
	{\centering
		\resizebox{1\linewidth}{!}{\includegraphics[trim={0cm 0cm 0cm 0cm},clip]{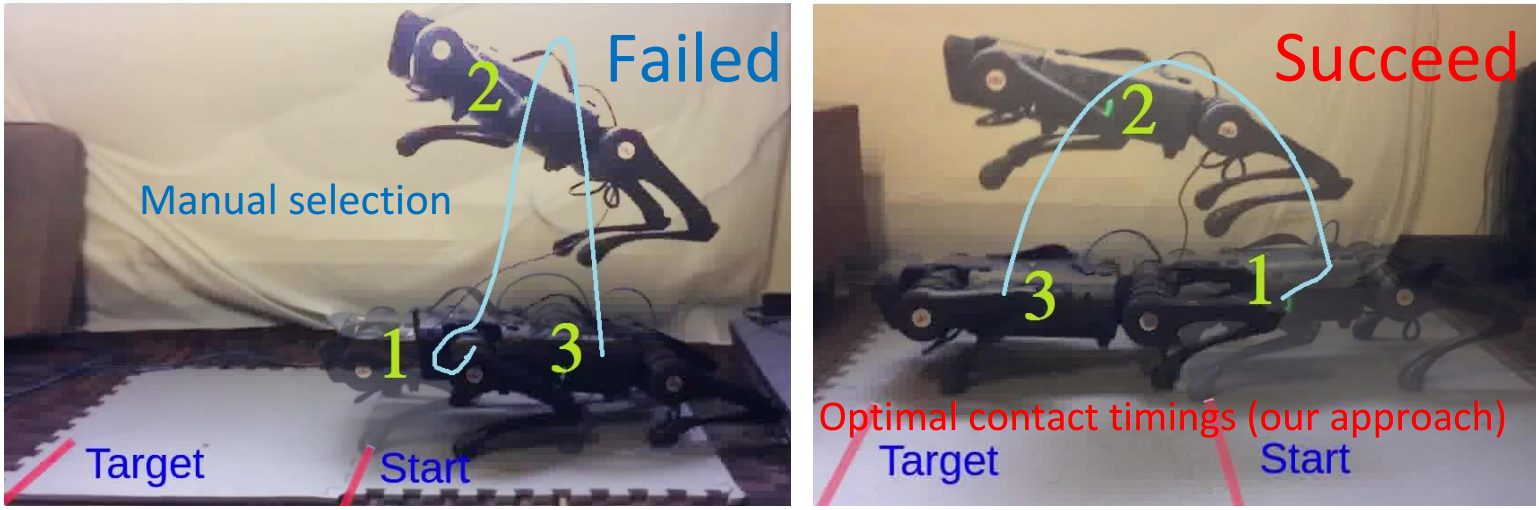}}
	}
	\caption{\textbf{Experiments:} Motion snapshots from jumping forward with manual selection of contact timings in the left figure and optimal contact timings (our approach) in the right figure.}
	\label{fig:timing_comparison_snapshot}
\end{figure}
\begin{figure*}[!t]
	\centering
	\subfloat[Trajectory comparison]{\centering
		\resizebox{0.29\linewidth}{!}{\includegraphics[trim={0cm 10cm 1cm 10cm},clip]{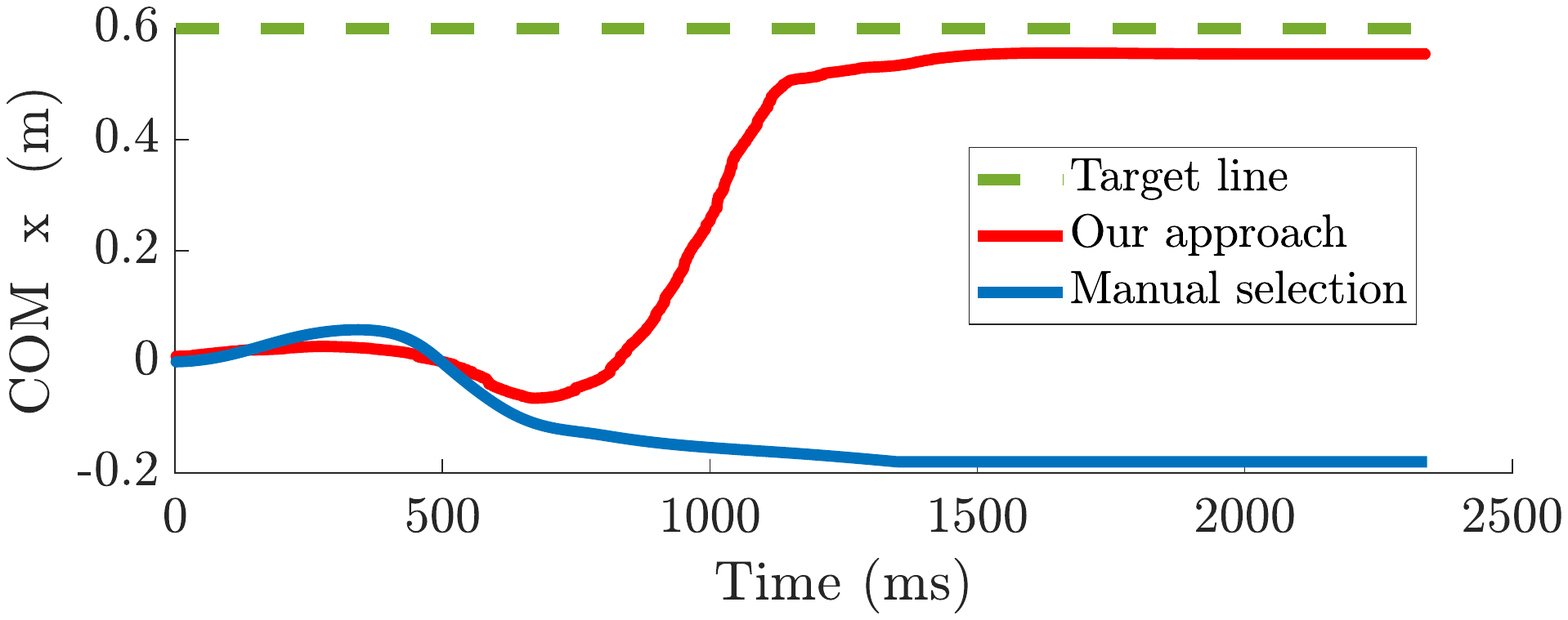}}
	}
	\subfloat[Thigh torque of rear legs]{\centering
		\resizebox{0.29\linewidth}{!}{\includegraphics[trim={1cm 10cm 1cm 10cm},clip]{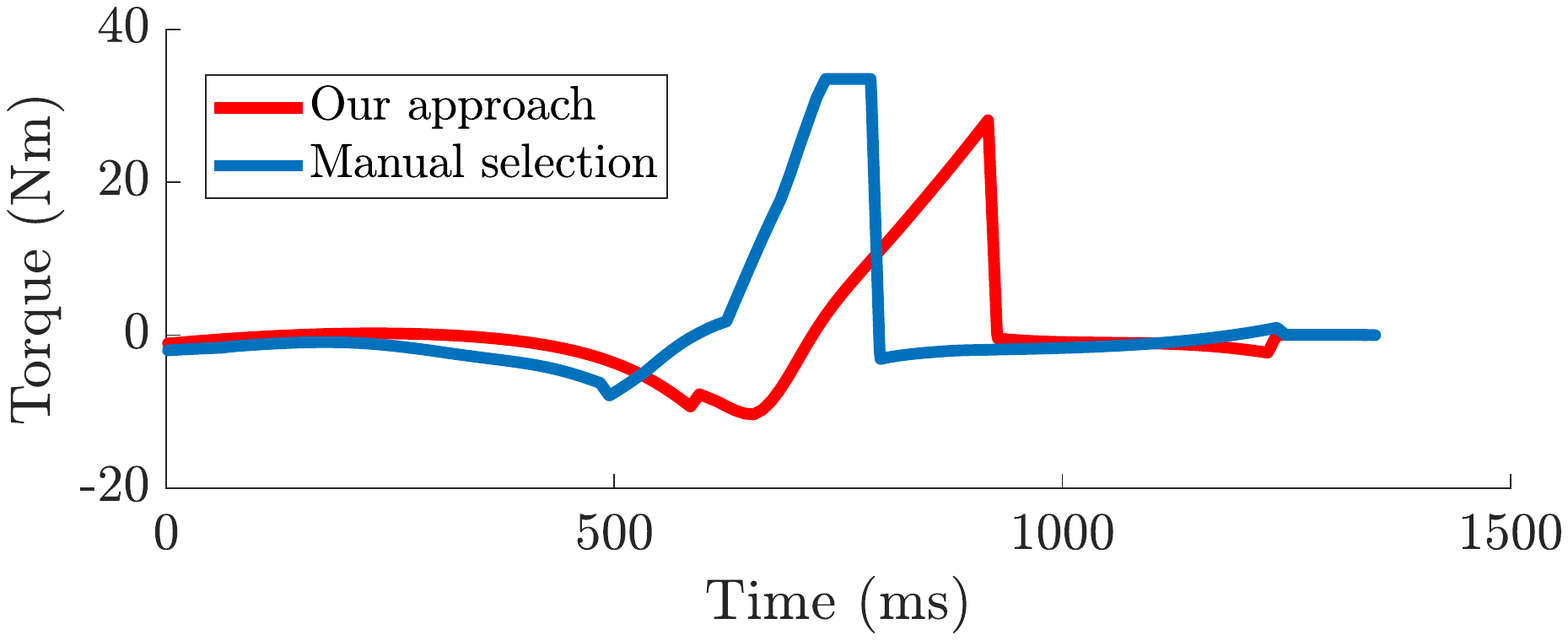}}
	}
	\subfloat[Calf torque of rear legs]{\centering
		\resizebox{0.29\linewidth}{!}{\includegraphics[trim={0cm 10cm 1cm 10cm},clip]{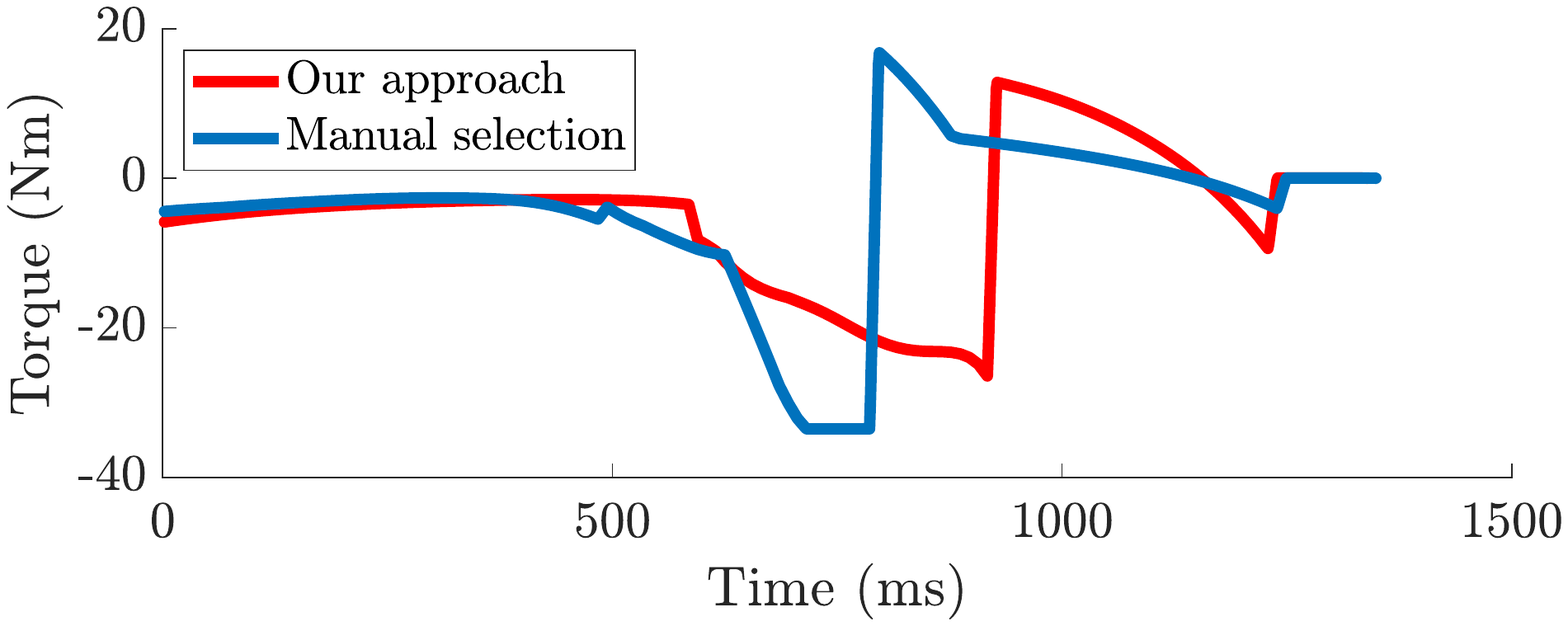}}
	}\\

	\caption{\textbf{Comparison between manual selection and optimal contact timings (our approach)} The comparison is considered based on target achievement from experiments in Fig. \ref{fig:timing_comparison}a, and the feed forward torque of rear legs generated from the full-body TO for jumping forward $0.6~m$ under manual selection of contact timings and optimal contact timing (our approach) in Fig. \ref{fig:timing_comparison}b \& \ref{fig:timing_comparison}c. The Fig. \ref{fig:timing_comparison} is accompanied with the Fig. \ref{fig:timing_comparison_snapshot}.}
	\label{fig:timing_comparison}
\end{figure*}
% \textcolor{blue}{Our results show that by efficiently implementing the TO for the full nonlinear dynamics model, we obtain highly accurate 3D aggressive jumping with only utilizing joint PD and Cartesian PD controller to track the reference, without the need to use additional controller such as WBC that is required to apply to compensate the inevitable error due to the use of simplified dynamics.}
Having introduced the full-body optimization approach, in this section we present a jumping and landing controller to help robot tracking the reference trajectory and effectively handle the high impact with the ground when performing 3D aggressive jumps off from high altitude (see Fig.\ref{fig:control_diagram}).
The desired joint angle $\bm{q}_d$, joint velocity $\bm{\dot q}_d$, foot position $\bm{p}_{f,d}$ and foot velocity $\bm{v}_{f,d}$ w.r.t their hips, and feed-forward joint torque $\bm{\tau}_d$ are obtained from the full-body TO. They are then linearly interpolated to get new reference profiles at $1~kHz$.  To track the reference trajectories, we use the feedback Cartesian PD controller that executes at $1~kHz$:
\begin{align}
\bm{\tau}_{ff}=\bm{J}(\bm{q}_{j})^\top[\bm{K}_{p}( \bm{p}_{f,d} - \bm{p}_d) +\bm{K}_{d}(\bm{v}_{f,d} - \bm{v}_f)] + \bm{\tau}_d \nonumber
\end{align} \label{eq:cartesian_controller}
where $\bm{J}(\bm{q}_{j})$ is the foot Jacobian at $\bm{q}_{j}$; $\bm{K}_p$ and $\bm{K}_d$ are diagonal matrices of proportional and derivative gains. The joint PD controller running at $10~kHz$ in the low-level motor control is integrated to improve the tracking performance.
The full controller for tracking desired trajectories is:
\begin{align}
\bm{\tau} = \bm{\tau}_{ff}+ \bm{K}_{p,joint}(\bm{q}_{j,d}-\bm{q})+\bm{K}_{d,joint}(\bm{\dot q}_{j,d}-\bm{\dot q}).  \label{eq:fullController}
\end{align}
Since there always exists a model mismatch between the optimization and hardware, it normally has orientation angle errors upon landing. Therefore, we utilize a real-time landing controller to handle impact, control GRF, balance the whole body motion during the landing phase, and recover the robot from unexpected landing configurations. For that controller, we extend the proposed QP controller \cite{QuannICRA19} for the 3D jumps:
\begin{align*}
&\textbf{if}~~ (\text{any}~ c_s >= \delta) \& (t\ge T_{posing}) \\
&\textbf{then}~~~\text{switch to landing controller},
\end{align*}
where $c_s$ is the value obtained from the contact sensor of the foot $s^{th}$, and $\delta$ is the force threshold to determine if the ground impact happened.
$T_{posing}$ is the instant at the beginning of the pre-landing configuration (after that instant $\dot{\bm{q}}_j^d=0$). When the impact is detected in any foot, we switch from jumping to landing controller, then based on which foots are in contact, different robot models are used for the landing controller. Our experimental results validate that it is effective to use the QP landing controller for SRB to handle impact with the ground and balance the robot when landing. 

\begin{remark}
Normally, all legs do not touch ground simultaneously due to the mismatch between the optimization model and hardware. Hence, utilizing a controller for swing legs also plays a crucial role here. Based on the contact model derived from the contact detection, the swing legs during landing phase are set at zero normal force and kept at the pre-landing configuration using PD controller until the ground contact is detected on these legs to prevent excessive impact force and unnecessarily extended movement.
\end{remark}
\begin{remark}
For highly aggressive jumping motions, e.g. double barrel roll and double backflips, due to the high linear and angular velocity of robot body upon the ground impact, the robot continues to move and rotate in the current direction, which may cause unfavorable landing posture as observed in our experiments. To tackle this issue, we utilize a PD controller during the pre-landing configuration to extend the legs in that direction \textcolor{black}{based on the body velocity reference obtained from trajectory optimization at the impact}.
% For example, for double barrel roll, the favorable foot position at the pre-landing is computed by
\end{remark}

%% file: Sections/Results.tex
\begin{comment}
\begin{table}[t!]
%	\hspace{0.2cm}
	\centering
	\caption{Solving time for whole-body TO}
	\label{tab:tab:3D_solving_time}
	\begin{tabular}{cccc}
		\hline
		Jumping tasks & Solving time (ipopt)\\
		\hline
		lateral jump 30cm &  31 [s] \\[.5ex]
		lateral jump down &  32 [s] \\[.5ex]
		$90^{o}$ spinning jump & 39 [s] \\[.5ex]
		diagonal jump &  226 [s]  \\[.5ex]
		barrel roll &  35 [s] \\[.5ex]
		double barrel roll &  46 [s] \\[.5ex]
		double backflip & 89 [s] \\[.5ex]
		
		\hline 
		\label{tab:3D_solving_time}
	\end{tabular}
\end{table}
\end{comment}

\section{Results}
\label{sec:Results}
This section presents experimental testing and
simulation results with the commercial Unitree A1 robot  \cite{A1}. A video of the results is included as supplementary material.  
\begin{comment}
In the cost function Eq.(\ref{eq:cost_function_wholebodyTO}),
We use a weight of $\epsilon_{\tau}= 0.003$ (while $\epsilon_q=1$, $\epsilon_N=1$).
\end{comment}
\begin{figure}[!t]
	\centering
	{\centering
		\resizebox{1.12\linewidth}{!}{\includegraphics[trim={2cm 10cm 0cm 5cm},clip]{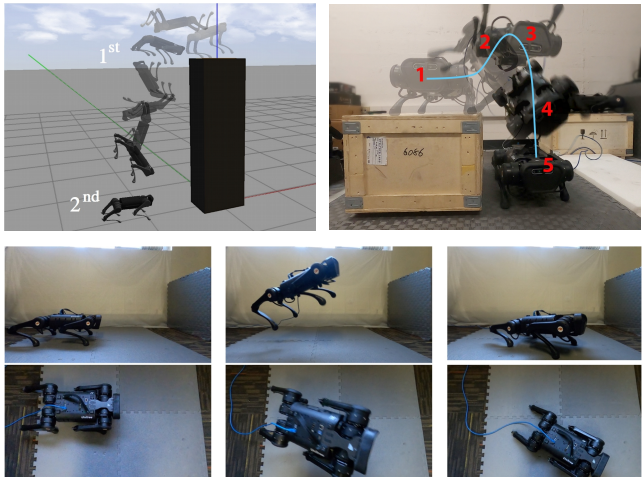}}
	}
	\\
	\caption{\textbf{Other example of successful 3D jumping motions achieved via our proposed framework.} The top left figure is a 3D \textit{double backflip} from a box of height $2m$. The top right is a 3D \textit{barrel roll} from a box of height $0.4m$. The figures in the last two rows are motion snapshots from a successful 3D \textit{diagonal jump} $[x,y,yaw]=[0.4~m, -0.3~m, 45^{0} ]$ from a side and a top view in experiments. } 
	\label{fig:3D_jumps}
\end{figure}

\begin{figure*}[!t]
	\centering
	\subfloat{\centering
		\resizebox{0.28\linewidth}{!}{\includegraphics[trim={0cm 10cm 1cm 10cm},clip]{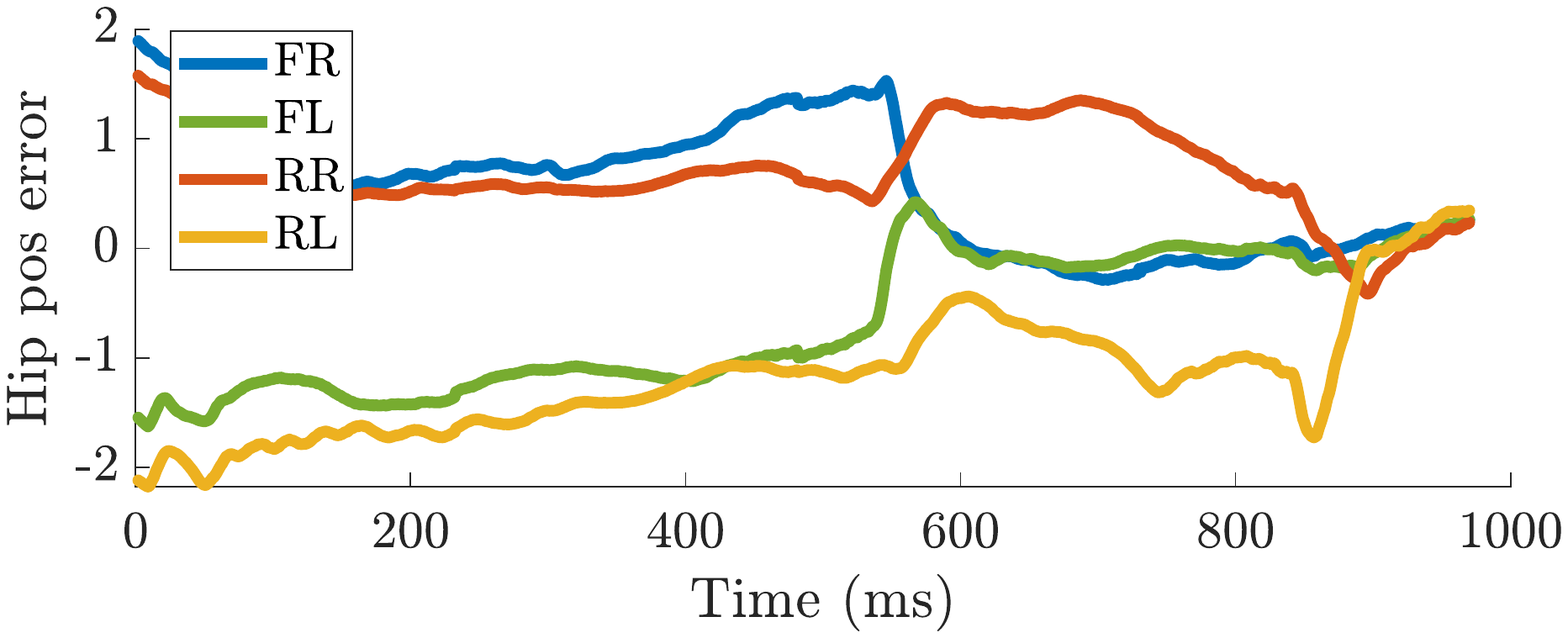}}
	}
	\subfloat{\centering
		\resizebox{0.28\linewidth}{!}{\includegraphics[trim={1cm 10cm 1cm 10cm},clip]{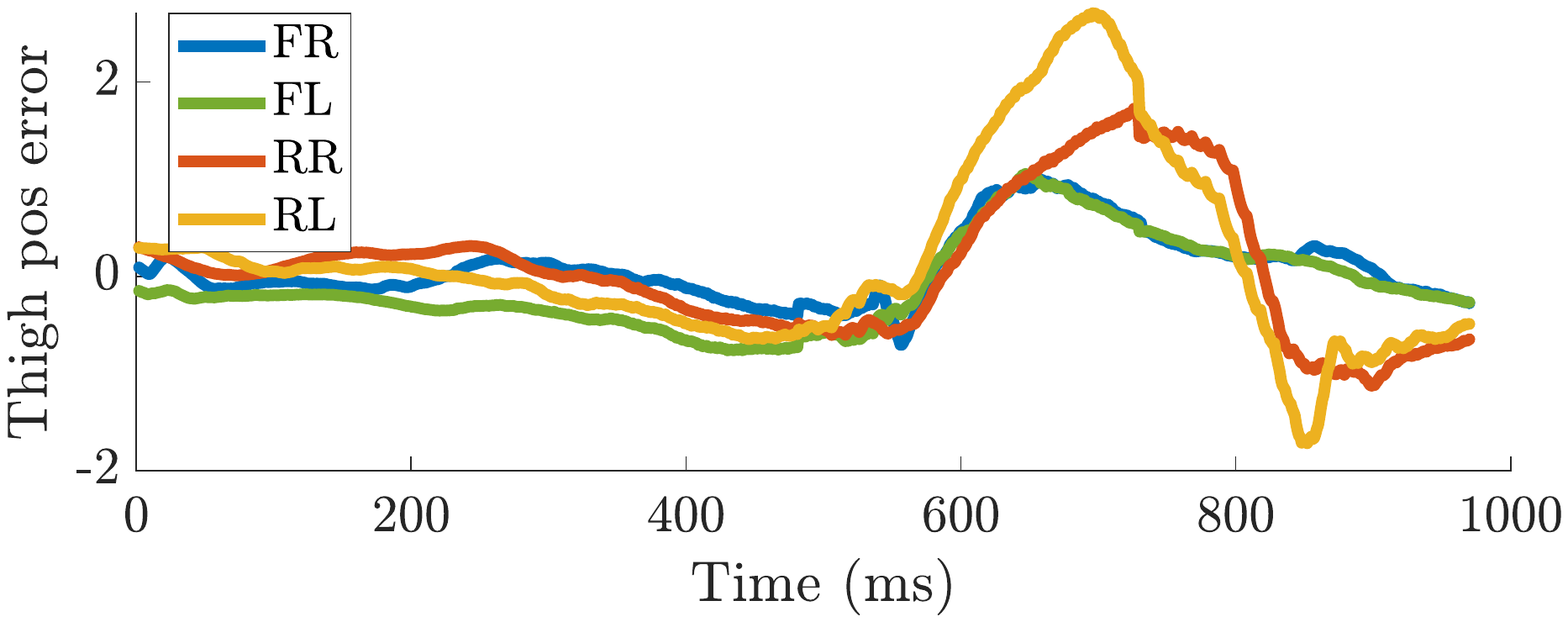}}
	}
	\subfloat{\centering
		\resizebox{0.28\linewidth}{!}{\includegraphics[trim={0cm 10cm 1cm 10cm},clip]{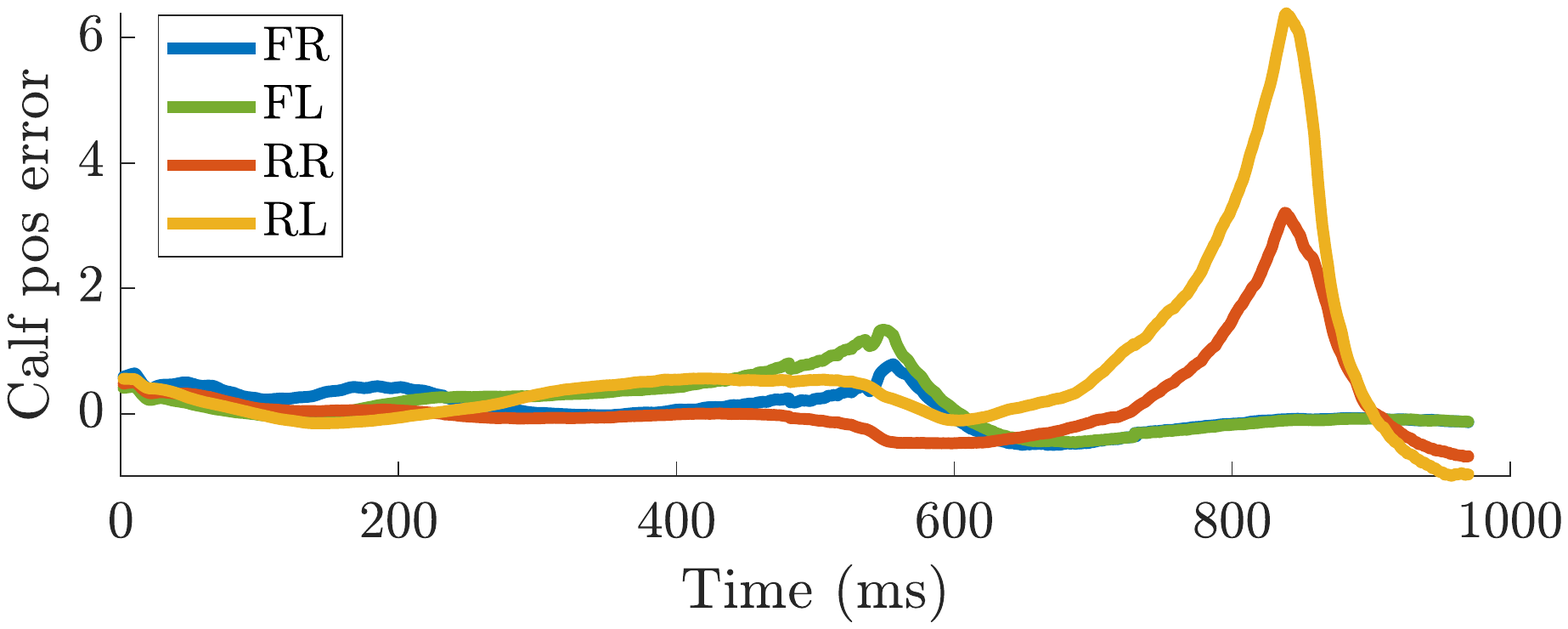}}
	}\\
% 	\subfloat{\centering
% 		\resizebox{0.28\linewidth}{!}{\includegraphics[trim={0cm 10cm 1cm 10cm},clip]{Figures/diagonal/COMe.pdf}}
% 	}
% 	\subfloat{\centering
% 		\resizebox{0.28\linewidth}{!}{\includegraphics[trim={1cm 10cm 1cm 10cm},clip]{Figures/diagonal/orientatione.pdf}}
% 	}
% 	\subfloat{\centering
% 		\resizebox{0.28\linewidth}{!}{\includegraphics[trim={0cm 10cm 1cm 10cm},clip]{Figures/diagonal/GRFe_2.pdf}}
% 	}\\

	\caption{\textbf{3D diagonal jump in experiments.} The figures are the tracking error of joint positions (degree) with jumping controller.
	FR, FL, RR and RL are used to denote front right, front leg, rear right, and rear left leg respectively.} 
	\label{fig:DiagonalJump_plots}
\end{figure*}

\begin{figure*}[!t]
	\centering
	\subfloat{\centering
		\resizebox{0.28\linewidth}{!}{\includegraphics[trim={0cm 10cm 1cm 10cm},clip]{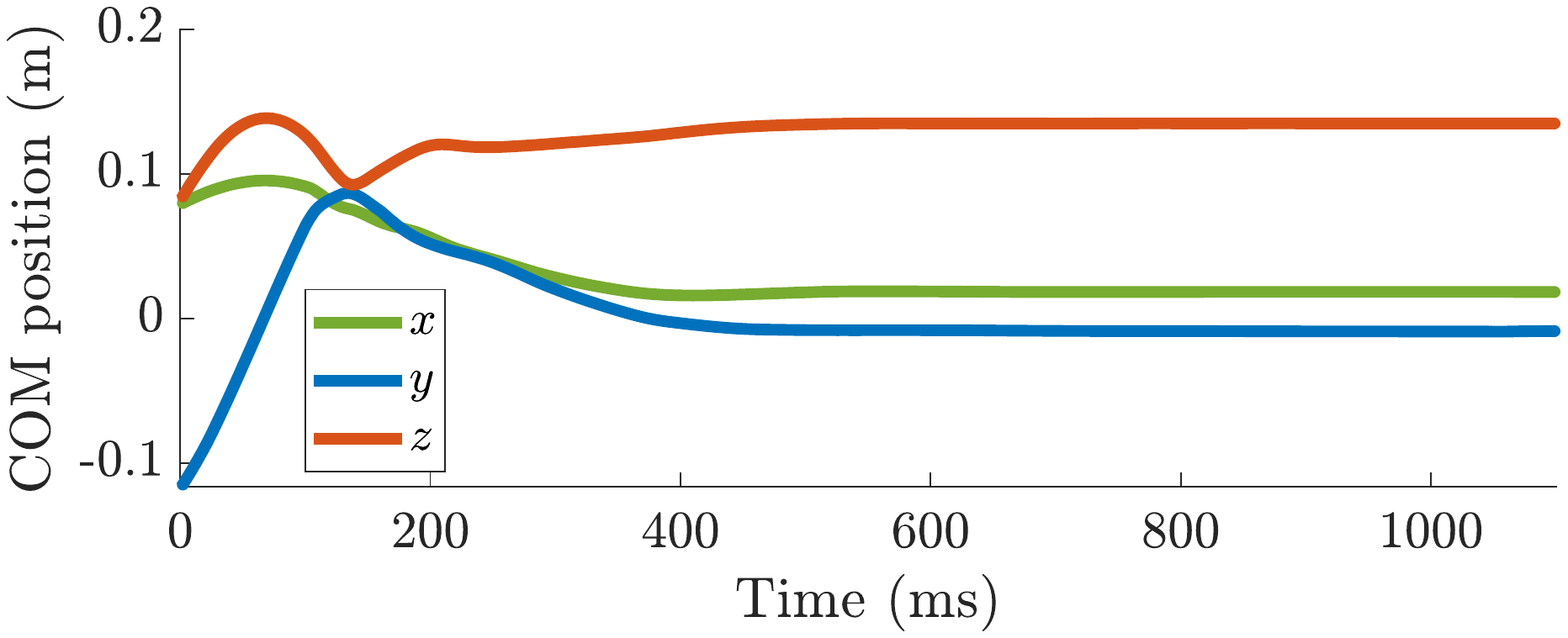}}
	}
	\subfloat{\centering
		\resizebox{0.28\linewidth}{!}{\includegraphics[trim={1cm 10cm 1cm 10cm},clip]{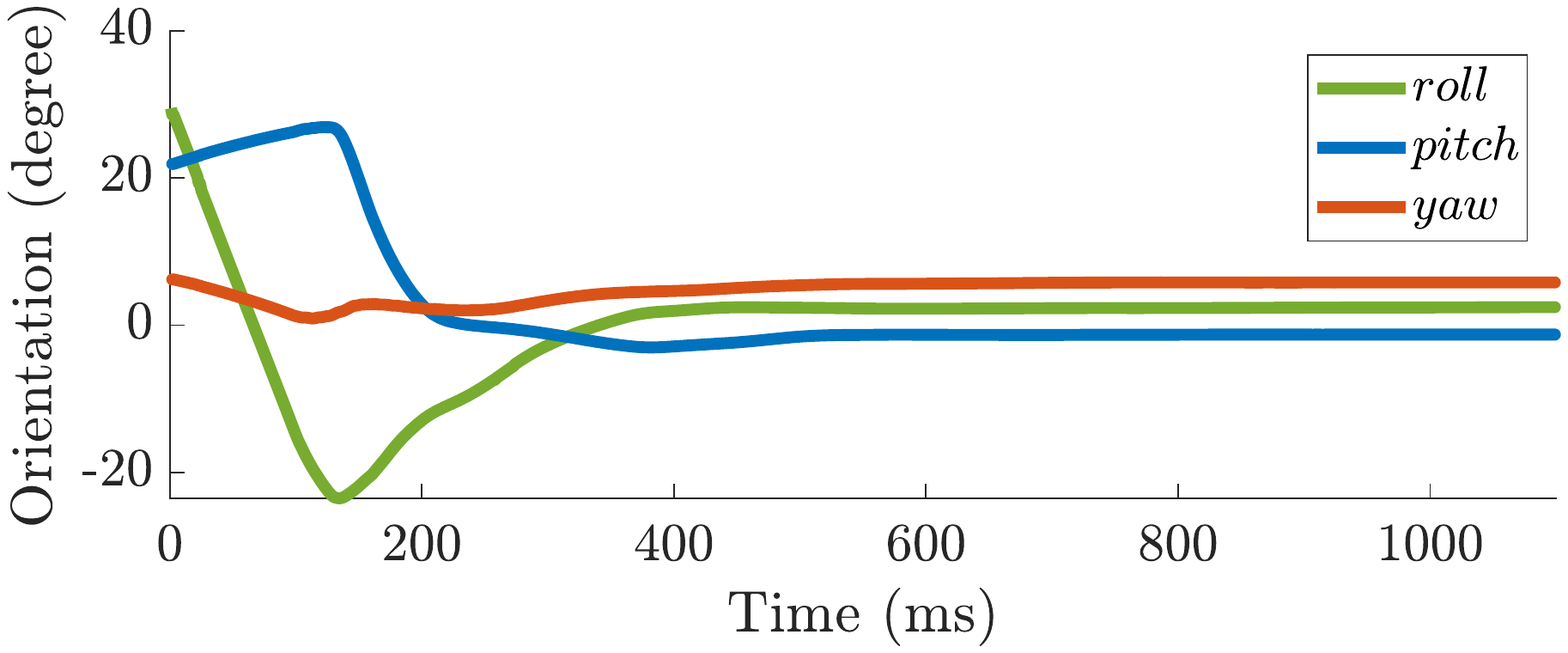}}
	}
	\subfloat{\centering
		\resizebox{0.28\linewidth}{!}{\includegraphics[trim={0cm 10cm 1cm 10cm},clip]{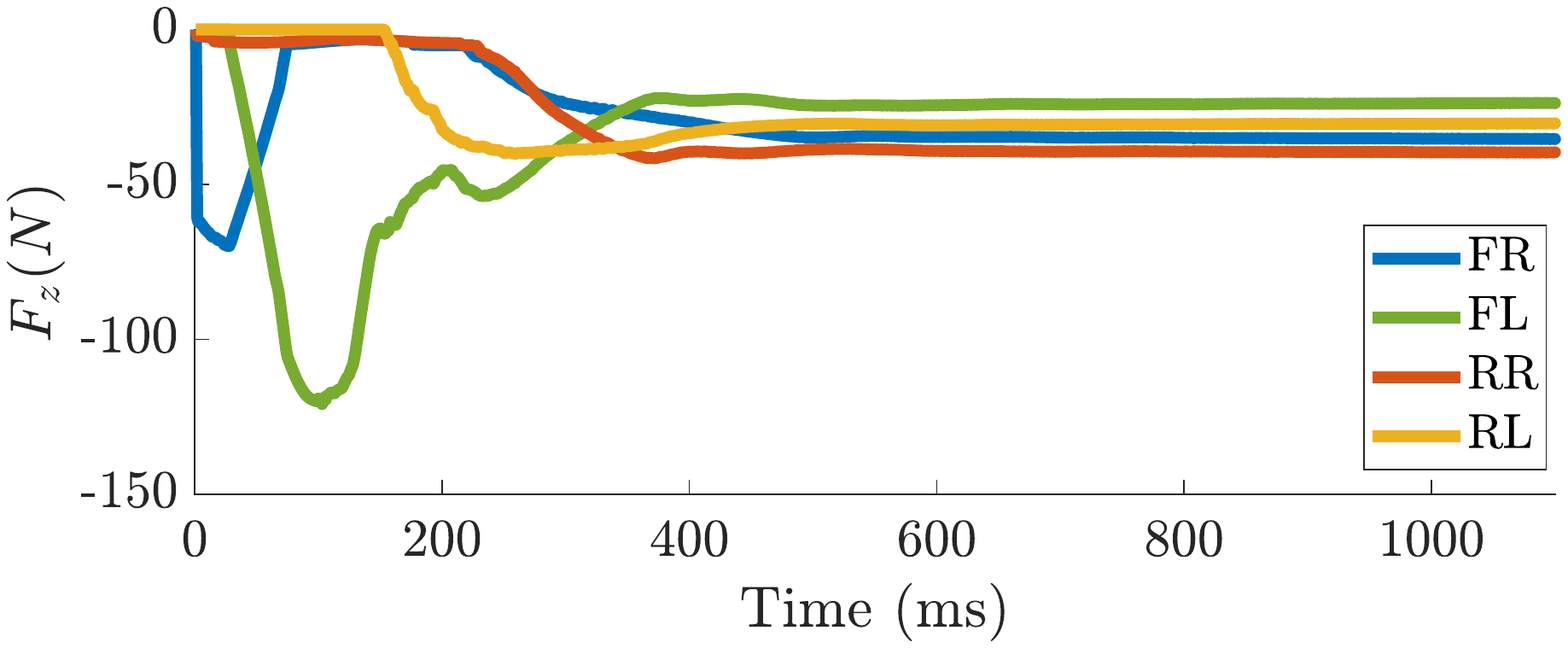}}
	}\\

	\caption{\textbf{Landing controller for a double barrel roll jump in experiments}. The plots show the COM position, body orientation, and force command during the landing phase when the robot performs a double barrel roll from a box with the height of $0.9~m$ (see Fig. \ref{fig:double_roll}).} 
	\label{fig:BarrelJump_plots}
\end{figure*}

\subsection{Numerical Simulation}
We use MATLAB and CasADi (see \cite{Andersson2018}) to construct and solve all presented trajectory optimization approaches for all 3D jumps. A user can predefine a contact sequence for the motion. \update{For example, for the lateral and spinning jumps, we use a sequence of a four-leg contact and then a flight phase.
For the total duration of the motion, we set $[T_{min}, T_{max}] = [0.5, 1.5] (s)$ (see \eqref{eq:time_opt_7}).  
With other jumping motions, the contact schedule is defined by a four-leg contact, a two-leg contact, and a flight phase, in which $[T_{min}, T_{max}] = [1, 2] (s)$. For all jumps, we use $N_i=30$ for each contact phase. As shown in Table \ref{tab:TO_solving_time}, our approach computes optimal contact timings given a wide range of $[T_{min}, T_{max}]$. It executes fast for all 3D jumps even with the high complexity of contact-timing optimization, which utilizes the rotation matrix representation}. In addition, the full-body trajectory optimization is solved at 
a tractable time (less than 4 minutes) for all jumping motions given the high complexity of the full-body dynamics.

\subsection{Experimental Validation}
% It has high torque density electric motors with single-stage 9:1 planetary gear reduction, and uses these actuators for all the hip, thigh, and knee joints to enable full 3D control of GRF. A pressure-based contact sensor is equipped on each foot. The A1 legs feature a large range of motion: the hip joints have a range of motion of $\pm$46${}^{\circ}$, the thigh joints have a range of motion from $- 60{}^ {\circ}$  to $ 240{}^ {\circ}$ and the knee joints have a range from $- 154.5{}^ {\circ}$ to $- 52.5{}^ {\circ}$. The hip and knee designs allow the robot to operate identically forward, backward and flipped upside-down. The lower link is driven by a bar linkage which passes through the upper link. The robot parameters and its actuation capabilities are summarized in Table \ref{tab:PRP}. In the next sections, we present the development of the framework.

Firstly, \update{in order to emphasize the importance of using full nonlinear dynamics model}, we implement the trajectory optimization of SRBD + full-body kinematic constraints to compare with our proposed full-body trajectory optimization. 
% The first TO approach solves for optimal robot body motion and joint references using SRBD with constraints on joint configuration \cite{matthew_mit2021_1}. 
% The first TO approach considers limits on joint configuration and joint torque \cite{matthew_mit2021_1}. Likewise \cite{matthew_mit2021_1}, joint torque is not an optimization variable.
In the first TO approach, constraints on joint torque $\bm{\tau}$ are imposed via joint angle $\bm{q}_j$ and ground reaction force $\bm{F}$ as the following simplification \cite{matthew_mit2021_1}: $\bm{\tau}_{min} \leq \bm{\tau}= J(\bm{q}_j)^T \bm{F} \leq \bm{\tau}_{max}$, where $J(\bm{q}_j)$ is the Jacobian of a leg. It is noted that in the first approach, the joint torque is not an optimization variable (similar to \cite{matthew_mit2021_1}).
% The first TO approach solves for the optimal robot body motion and joint references while considering constraints on joint configuration and joint torque \cite{matthew_mit2021_1}. Moreover, it is noted that in the first approach, the joint torque is not an optimization variable (similar to \cite{matthew_mit2021_1}).
% , but we use joint space instead of Cartesian space to enforce related kinematic constraints. 
% \update{We also ignore (\ref{eq:time_opt_7}), and set $T_i/N_i =dt$ in (\ref{eq:time_opt_6}) as a time step for all phases, e.g., $dt=10ms$}. 
% \todo{explain briefly about full kinematic constraints}.
% It is noted that in the first approach, the joint torque is not an optimization variable (similar to \cite{matthew_mit2021_1}).
For the comparison, we pick a $90^{0}$ spinning jump with the contact schedule of (56,31) computed in Table \ref{tab:TO_solving_time}, to discuss here. As shown in Fig. \ref{fig:yaw_compare}, while our approach using the full-body dynamics guarantees a highly accurate jump, the trajectory optimization of SRBD and kinematic constraints fails to achieve the target angle. 
\update{It has shown that the legs’ dynamics, which is neglected in the simplified dynamics, has a significant contribution in highly dynamic motions. Note that the comparison is made by directly applying joint PD controllers  \eqref{eq:fullController} to track the desired joint profiles obtained from these trajectory optimization approaches.} 
% instead of utilizing other high level tracking controllers (e.g., WBC \cite{GabrielICRA2021} or Variational-Based Optimal Control \cite{ChignoliICRA2021}) 
% \todo{why mention other approach here?. 

% \todo{failure of what, need to be more specific and make connection with previous sentences. "it" is also not clear, need to be more specific. }

Secondly, \update{we compare optimal contact timings with manually selected contact timings to show the advantage of the contact-timing optimization}. Experiments for jumping forward $0.6~m$ are picked to discuss here (see Fig. \ref{fig:timing_comparison_snapshot} and Fig. \ref{fig:timing_comparison}). If timings are manually selected with unnecessary long flight time (e.g., $550~ms$), this makes the motor's torque saturated in $100~ms$ that is up to $1/3$ of rear-leg contact phase. This selection seriously affects the motor working condition, causing failed joint tracking performance. Thus, the robot is unable to reach the target. On the other hand, selecting too small flight time makes the optimization unsolvable since the robot does not have enough power to jump to the desired configuration. By using optimal contact timings, we minimize the effort or energy, prevent the torque saturation issue, and execute successful jumps on the hardware.
\begin{comment}
\begin{figure}[!t]
	\centering
	{\centering
		\resizebox{0.8\linewidth}{!}{\includegraphics[trim={0cm 10cm 0cm 10cm},clip]{Figures/SRB_and_whole_body_TO/yaw_compare2.pdf}}
	}
	\caption{\textbf{} Comparison between and TO of SRB+ full kinematics approach and whole-body TO (our approach) for $90^0$ spinning jump target.}
	\label{fig:yaw_compare}
\end{figure}
\end{comment}

Finally, we present the results of different aggressive
3D jumping experiments achieved by our framework that combines
optimal contact timings and full-body TO. 
To the best of our knowledge, the results herein is the first implementation
of full-body TO of quadruped robots for 3D jumping motions. 
We pick some jumping tasks to discuss here.
In a successful diagonal jump (see Fig. \ref{fig:3D_jumps}), the jumping controller guarantees the high tracking performance as evident in Fig. \ref{fig:DiagonalJump_plots}. 
Using our approach, the robot can successfully performs a barrel roll from a box of height $0.4~m$ in experiments (see Fig. \ref{fig:3D_jumps}). Especially, with our proposed framework, the robot is able to complete a highly aggressive gymnastics double barrel roll jump, which is the first jump ever achieved by the quadrupedal robots (see Fig.\ref{fig:double_roll}). 
% , during the jumping phase with left-leg contact, the system is under-actuated and feedback control is not utilized on the body orientation. 
These results illustrates the efficiency of our approach in optimizing the contact timings for the jumping task as well as the accuracy of the optimization framework when realizing in the robot hardware.
\update{In addition, these results also validate the effectiveness of the landing controller in handling hard impact.} For that double barrel roll experiment, since the robot rotates with a high angular velocity from a high altitude, it has a considerably hard impact with the ground. However, the landing controller can recover the robot's body position and orientation under the hard impact and significant error of landing configuration (see Fig. \ref{fig:BarrelJump_plots}).

%% file: Sections/Conclusion.tex
%!TEX root = ../Main.tex

\section{Conclusions}
\label{sec:Conclusion}
\textcolor{black}{This paper has introduced the framework for performing highly dynamics 3D jumps with long aerial time on quadruped robots that require model accuracy, perfect timings and coordination of the whole body motion. This framework combines contact timings of SRB dynamics, full-body TO, the jumping controller, and robust landing controller. The efficiency of the framework is validated via both A1 robot model and experiments on performing these aggressive tasks. 
The vision to autonomously detect obstacles will be integrated with the proposed framework in our future work.
}
\section*{Acknowledgments}
The authors would like to thank Hiep Hoang and Yiyu Chen at Dynamic Robotics and Control lab for their assistance in simulation and hardware experimentation. We also thank Dr. Roy Featherstone for the insightful discussion about Spatial v2 for rigid body dynamics.